\documentclass[conference]{IEEEtran}
\usepackage{cite}
\usepackage{amsmath,amssymb,amsfonts}
\usepackage{algorithmic}
\usepackage{graphicx}
\usepackage{textcomp}
\usepackage{xcolor}
\usepackage{booktabs, multirow} 
\usepackage{soul}
\usepackage{subcaption}
\usepackage{xcolor,colortbl} 
\def\BibTeX{{\rm B\kern-.05em{\sc i\kern-.025em b}\kern-.08em
    T\kern-.1667em\lower.7ex\hbox{E}\kern-.125emX}}

\makeatletter
\def\endthebibliography{%
  \def\@noitemerr{\@latex@warning{Empty `thebibliography' environment}}%
  \endlist
}
\makeatother

\usepackage{amssymb}
\usepackage{pifont}

\definecolor{Green}{rgb}{0,0.7,0}

\IEEEoverridecommandlockouts

\begin{document}

\makeatletter
\newcommand{\linebreakand}{%
  \end{@IEEEauthorhalign}
  \hfill\mbox{}\par
  \mbox{}\hfill\begin{@IEEEauthorhalign}
}
\makeatother

\title{Learning based Ge'ez character handwritten recognition}

\author{\IEEEauthorblockN{Hailemicael Lulseged Yimer}
\IEEEauthorblockA{\textit{Department of Engineering DIMI} \\
\textit{University of Verona}\\
Verona, Italy \\
hailemicaellulseged.yimer@studenti.univr.it}
\and
\IEEEauthorblockN{Hailegabriel Dereje Degefa}
\IEEEauthorblockA{\textit{Department of Engineering DIMI} \\
\textit{University of Verona}\\
Verona, Italy \\
hailegabrieldereje.degefa@studenti.univr.it}
\linebreakand
\IEEEauthorblockN{Marco Cristani}
\IEEEauthorblockA{\textit{Department of Engineering DIMI} \\
\textit{University of Verona}\\
Verona, Italy \\
marco.cristani@univr.it}
\and
\IEEEauthorblockN{Federico Cunico}
\IEEEauthorblockA{\textit{Department of Engineering DIMI} \\
\textit{University of Verona}\\
Verona, Italy \\
federico.cunico@univr.it}
\thanks{This study was carried out within the PNRR research activities of the consortium iNEST (Interconnected North-Est Innovation Ecosystem) funded by the European Union Next-GenerationEU (Piano Nazionale di Ripresa e Resilienza (PNRR) – Missione 4 Componente 2, Investimento 1.5 – D.D. 1058 23/06/2022, ECS\_00000043). This manuscript reflects only the Authors’ views and opinions. Neither the European Union nor the European Commission can be considered responsible for them.}
}

\IEEEoverridecommandlockouts
\IEEEpubid{\makebox[\columnwidth]{979-8-3503-7838-2/24/\$31.00~\copyright2024 IEEE \hfill} \hspace{\columnsep}\makebox[\columnwidth]{ }}

\maketitle
\IEEEpubidadjcol

\begin{abstract}

Ge'ez, an ancient Ethiopic script of cultural and historical significance, has been largely neglected in handwriting recognition research, hindering the digitization of valuable manuscripts. Our study addresses this gap by developing a state-of-the-art Ge'ez handwriting recognition system using Convolutional Neural Networks (CNNs) and Long Short-Term Memory (LSTM) networks. Our approach uses a two-stage recognition process. First, a CNN is trained to recognize individual characters, which then acts as a feature extractor for an LSTM-based system for word recognition. Our dual-stage recognition approach achieves new top scores in Ge'ez handwriting recognition, outperforming eight state-of-the-art methods, which are SVTR, ASTER,and others as well as human performance, as measured in the HHD-Ethiopic dataset work.
This research significantly advances the preservation and accessibility of Ge'ez cultural heritage, with implications for historical document digitization, educational tools, and cultural preservation.
The code will be released upon acceptance.
\end{abstract}

\section{Introduction}

Handwritten recognition faces challenges due to varied writing styles and character overlaps. Intelligent systems using classification methods are effective in categorizing handwritten characters as humans do~\cite{plamondon2000online}. Deep learning algorithms, particularly Convolutional Neural Networks (CNNs), have shown exceptional performance by automatically extracting features from raw images~\cite{lecun1998gradient}. CNNs are suitable for handwritten recognition due to their ability to learn directly from input data~\cite{moges2017study}.
CNNs have become dominant in handwritten character recognition, excelling at capturing spatial hierarchies and local features in images~\cite{krizhevsky2017imagenet}. However, they are limited to processing spatial information without considering sequential context. This limitation highlights the potential benefits of integrating Recurrent Neural Networks (RNNs), especially Long Short-Term Memory (LSTM) networks~\cite{hochreiter1997long}.

LSTMs are designed to handle sequential data and capture temporal dependencies, making them suitable for tasks where character context and order are crucial. A hybrid model combining CNNs and LSTMs can leverage the strengths of both: CNNs for local feature extraction and LSTMs for sequential processing. This integrated approach can lead to more accurate and robust recognition of handwritten characters, addressing the variability and complexity in individual writing styles and sequences~\cite{wigington2017data}.

\begin{figure}[t]
    \centering
    \includegraphics[width=0.8\linewidth]{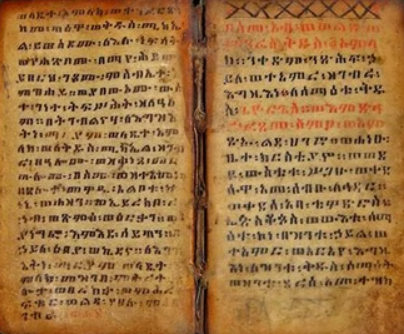}
    \caption{Example of a Ge'ez handwritten manuscript. This historical document exemplifies the challenges in optical character recognition (OCR) for low-resource scripts, highlighting the intricate and variable nature of Ge'ez handwriting. It underscores the necessity of advanced machine learning and deep learning techniques to accurately digitize and preserve such culturally significant texts}
    \label{fig:example}
\end{figure}

These challenges and approaches are particularly relevant when considering specific writing systems like Ge'ez~\cite{geez}, an archaic writing system widely employed in Ethiopia and Eritrea, which poses distinctive difficulties for digital preservation and recognition. There is a pressing need to digitize thousands of hand-written documents in Ge'ez to prevent their deterioration and ensure their long-term preservation. At present, there is a notable deficiency in efficient handwriting recognition algorithms specifically designed for the Ge'ez script. This deficiency poses a risk to the potential loss of priceless cultural and historical heritage. An example of Ge'ez document is visible in Figure~\ref{fig:example}.

In this work, we want to help this process by creating a sophisticated handwriting recognition system specifically designed for the Ge'ez script. 
We propose an innovative CNN-LSTM architecture for word-level recognition
where CNNs act as feature extractors feeding into LSTM-based sequence modeling. Our work leverages a comprehensive dataset of handwritten Ge'ez characters and words, enabling thorough training and evaluation of our recognition systems. 
Our contributions are the following:
\begin{itemize}
    \item We propose a novel method based on CNN and LSTM for handwritten Ge'ez scripts;
    \item WIth a Cer of 26.95 and a NED of 26.50, our model achieves state of the art performance for Ge'ez optical character recognition;
\end{itemize}

The paper is structured as follows: Section \ref{sec:sota} presents related works in the handwritten Ge'ez recognition, Section~\ref{sec:methodology} presents our method, Section~\ref{sec:experiments} show the efficacy of our method presenting experiments and discussing the results, in comparison with the state of the art, and finally in Section~\ref{sec:conclusions} we present the conclusions of the work.

\section{Related work}\label{sec:sota}

\subsection{Datasets}
    
The field of Ethiopic script character recognition, particularly for historical handwritten documents, faces a significant challenge due to the scarcity of available datasets. Before our work, the only dataset available in the literature was the HHD-Ethiopic dataset introduced by \cite{belay2023hhd}. This scarcity of data has been a major limiting factor in developing and evaluating OCR systems for Ethiopic script.
The HHD-Ethiopic dataset, while groundbreaking, represents just the beginning of the necessary resources for this field. 
Its introduction highlights the critical need for more comprehensive and diverse datasets to advance research in Ethiopic script recognition. The limited availability of annotated data underscores the importance of our contribution and its potential impact on future studies in this area.

\subsection{Handwritten character recognition}

In recent years, Optical Character Recognition systems (OCRs) have been increasingly explored for various scripts and languages. The most similar language to Ge'ez is Amharic, which shares a common script and linguistic roots. However, while Amharic OCR has seen significant advancements, the unique characteristics and lower resource availability of Ge'ez present distinct challenges that require specialized approaches in OCR development. 

In the work of ~\cite{belay2023hhd}, the authors proposed an end-to-end Amharic OCR system utilizing traditional CTC-based architectures like Plain-CTC and Attention-CTC, incorporating CNNs and Bi-LSTMs for feature extraction and sequence prediction, with attention mechanisms to enhance recognition of complex script nuances. Their approach was evaluated on the HHD-Ethiopic dataset, featuring historical handwritten text-line images from 18th to 20th-century Amharic and Ge’ez manuscripts, addressing the lack of resources for recognizing these complex scripts.  

Similarly, authors of ~\cite{belay2020amharic} introduced a blended Attention-CTC network architecture for Amharic text recognition, combining CNNs and Bidirectional LSTM for feature extraction and sequence learning, with attention mechanisms directly integrated within the CTC framework on the ADOCR dataset. Furthermore, It uses an encoder, attention, and transcription module in a unified framework. 

Belay et al.~\cite{belay2020ocr} proposed an end-to-end Amharic OCR approach using a neural network that combines a CNN-based feature extractor, Bidirectional LSTM for sequence learning, and a Connectionist Temporal Classification transcriber. The model, trained on the ADOCR dataset with various Amharic fonts, eliminates the need for character-level segmentation. 

Du et al.~\cite{du2022svtr} introduced SVTR, a single visual model for scene text recognition that integrates feature extraction and text recognition in a unified framework. Their approach involves decomposing text images into small patches, termed character components, which are processed through a three-stage architecture featuring global and local mixing blocks to capture both intra-character and inter-character dependencies.  

Shi et al.~\cite{shi2016end} proposed a Convolutional Recurrent Neural Network (CRNN) for scene text recognition, integrating feature extraction, sequence modeling, and transcription into a single end-to-end trainable framework. This approach uses convolutional layers for extracting sequential features, bidirectional LSTM layers for sequence learning, and a transcription layer with Connectionist Temporal Classification (CTC) for output prediction.

\begin{figure}[t]
    \centering
    \begin{subfigure}[b]{0.8\linewidth}
        \centering
        \includegraphics[width=0.45\linewidth]{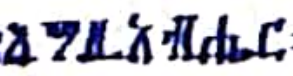}
        \includegraphics[width=0.15\linewidth]{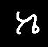}
        \caption{Ge'ez word and character in handwriting style 1}
    \end{subfigure}
    \vspace{1em}
    \begin{subfigure}[b]{0.8\linewidth}
        \centering
        \includegraphics[width=0.45\linewidth]{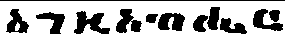}
        \includegraphics[width=0.15\linewidth]{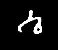}
        \caption{Same Ge'ez word and character in handwriting style 2}
    \end{subfigure}
    \vspace{1em}
    \begin{subfigure}[b]{0.8\linewidth}
        \centering
        \includegraphics[width=0.9\linewidth]{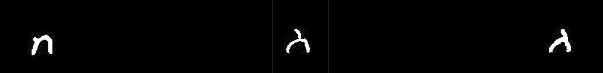}
        \caption{Subset of Ge'ez characters that are challenging to discriminate}
    \end{subfigure}
    \caption{Challenges in Ge'ez handwriting recognition. (a) and (b) demonstrate the variation in handwriting styles for the same word and characters. As visible, the difference in appearance is significant, highlighting the complexity of recognition across different scribes or historical periods. (c) shows a subset of the 182 Ge'ez characters, focusing on those that are particularly difficult to distinguish due to their similar forms.}
    \label{fig:styles}
\end{figure}

Shi et al.~\cite{shi2018aster} proposed ASTER, an attentional scene text recognizer with flexible rectification, designed to handle distorted or irregular text, such as perspective or curved text. ASTER combines a rectification network using Thin-Plate Spline (TPS) transformation to correct distortions and a recognition network that predicts character sequences using an attentional sequence-to-sequence model. It performs on irregular text recognition without needing extra annotations. 

Fang et al.~\cite{fang2021read} introduced ABINet, a text recognition method that models linguistic behavior in an autonomous, bidirectional, and iterative manner. The ABINet architecture decouples the vision model (VM) and language model (LM), enforcing explicit language modeling by blocking gradient flow between them. They utilized a bidirectional cloze network (BCN) to capture bidirectional context and employed an iterative correction mechanism to refine predictions progressively, enhancing recognition of low-quality images.

\section{Method}\label{sec:methodology}

\begin{figure}[t]
    \centering
    \begin{subfigure}[b]{\linewidth}
        \centering
        \includegraphics[width=\linewidth]{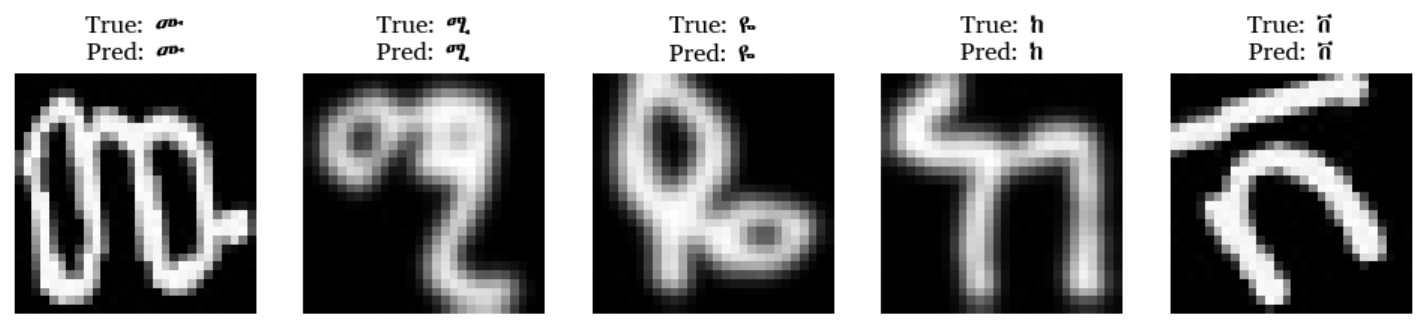} 
        \caption{}
        \label{fig:character_predictions_characters}
    \end{subfigure}
    \vspace{1em} 
    \begin{subfigure}[b]{\linewidth}
        \centering
        \includegraphics[width=\linewidth]{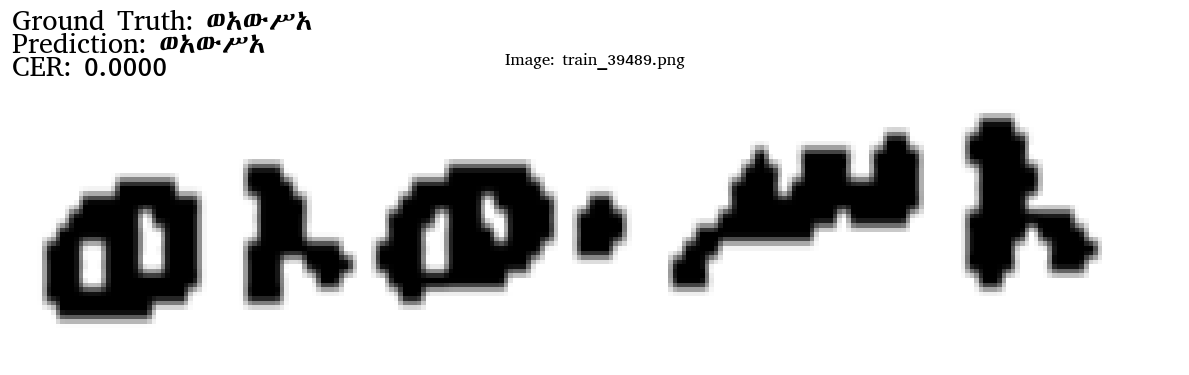} 
        \caption{}
        \label{fig:character_predictions_text}
    \end{subfigure}
    \caption{Examples of predictions made by the model. (a) Character-level predictions. (b) Text-level predictions.}
    \label{fig:merged_predictions_vertical}
\end{figure}

\subsection{Architecture}

The OCR model architecture integrates Convolutional Neural Networks (CNNs) and Long Short-Term Memory (LSTM) networks to recognize Amharic text in images. It begins with a series of convolutional layers organized into residual blocks for feature extraction. These blocks incorporate residual connections to facilitate better gradient flow in the deep network. Each block is followed by max pooling to reduce spatial dimensions and dropout for regularization. Batch normalization is applied after convolutional layers to stabilize training and improve convergence. The extracted features then feed into two stacked bidirectional LSTM layers, capturing sequential dependencies in both directions of the text. Dropout is also applied between LSTM layers to prevent overfitting. A final fully connected layer produces character probabilities for each time step. This hybrid architecture leverages CNNs' strength in visual feature extraction and LSTMs' proficiency in processing sequential data. 
The model is trained using Connectionist Temporal Classification (CTC) loss, enabling learning from unsegmented input sequences. This approach is particularly beneficial for handwritten text recognition where character boundaries are not explicitly defined. CTC aligns predicted character sequences with ground truth, handling variable-length outputs and allowing for blank tokens or repeated characters as needed. By combining these architectural elements with various regularization techniques, the model achieves a robust and generalized approach to the complex task of Amharic text recognition.

\subsection{Implementation details}
The proposed model architecture integrates CNN and LSTM networks to effectively process and recognize Ge'ez text in images. At its core, the model employs a series of four residual blocks, each comprising convolutional layers with ReLU activation and batch normalization, followed by max pooling and dropout. These blocks progressively increase the number of channels from 64 to 512, extracting hierarchical features while reducing spatial dimensions. The residual connections within these blocks facilitate better gradient flow, enabling easier training of the deep network. Following the convolutional layers, the extracted features are reshaped and fed into two stacked bidirectional LSTM layers, each with 512 units in both directions. These LSTMs capture sequential dependencies in both forward and backward directions, which is crucial for understanding the context of Ge'ez's text. A
Finally, a fully connected layer maps the attended features to character probabilities for each time step, corresponding to the Ge'ez character set, plus a blank token for CTC loss.

Also, the model was trained and evaluated using a Linux operating system and GPU Nvidia RTX 3090. The computational setup ensured the efficient handling of large datasets and provided the environment for deep learning model training and performance evaluation.




\section{Experiments and Results}\label{sec:experiments}


We use the HHD-Ethiopic dataset for handwritten text recognition comprises 79,684 samples, including two key test sets: an In-Distribution (IID) set of 6,375 samples drawn randomly from the training distribution, and an Out-of-Distribution (OOD) set of 15,935 samples from 18th-century manuscripts, with in the testing distribution. The IID/OOD split enables an evaluation of model performance on both familiar and historically challenging samples. We evaluate our model against several existing approaches, as detailed in Section~\ref{sec:sota}, with results summarized in Table~\ref{tab:sota}.
\begin{figure}[t]
    \centering
    \includegraphics[width=\linewidth]{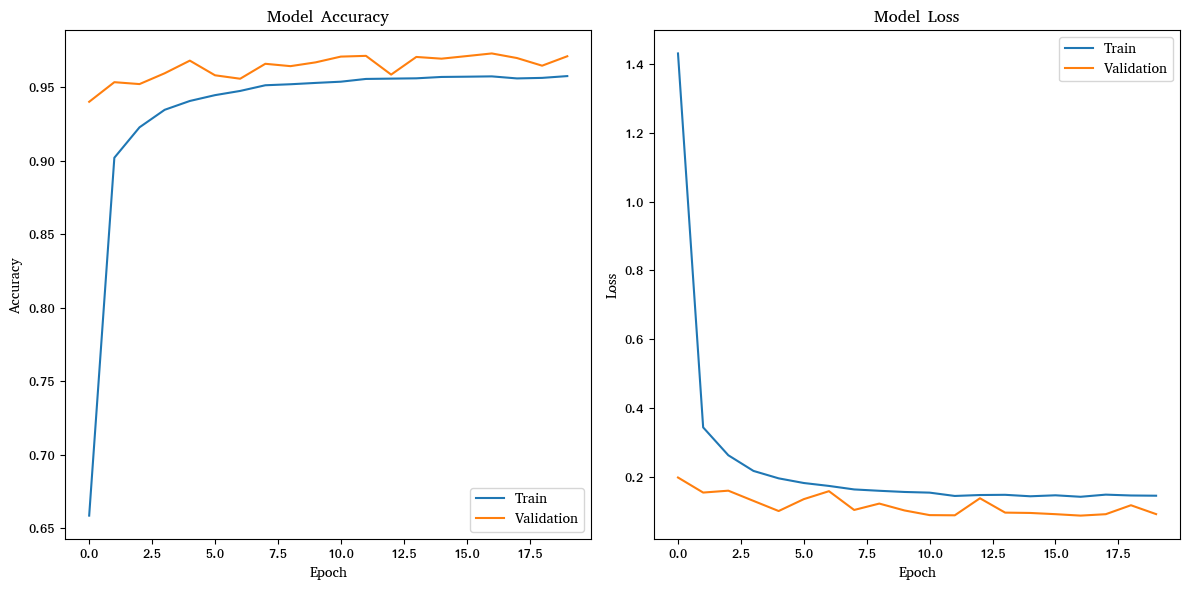} 
    \caption{The loss and classification accuracy of the character recognition model during training. The system seems to learn the task in a few epochs, and further training did not significantly improve accuracy.}
    \label{fig:loss}
\end{figure}
We employ two widely-used metrics in handwritten text recognition: Character Error Rate (CER) and Normalized Edit Distance (NED). Both metrics were also solely used in prior works, such as~\cite{belay2023hhd}, as standard benchmark for evaluating similar models.
Character Error Rate (CER) is a metric used to evaluate the performance of text recognition systems. It measures the edit distance between the predicted text and the ground truth text at the character level. The formal definition of CER is:
\begin{equation}
    CER = \frac{S + D + I}{N}
\end{equation}
Where S is the number of substitutions, D is the number of deletions, I is the number of insertions, and N is the total number of characters in the ground truth. CER is expressed as a percentage, with lower values indicating better performance.

Normalized Edit Distance (NED) is another metric used in text recognition evaluation. It's based on the Levenshtein distance (edit distance) between the predicted text and the ground truth, but it's normalized to account for varying text lengths. The formal definition of NED is:
\begin{equation}
 NED = \frac{EditDistance(\text{predicted}, \text{ground\_truth})}{max(len(\text{predicted}), len(\text{ground\_truth}))}   
\end{equation}

Where $EditDistance()$ calculates the minimum number of single-character edits (insertions, deletions, or substitutions) required to change one word into another. NED is expressed as a value between 0 and 1, with lower values indicating better performance.


\begin{table}[!htp]
\centering
\caption{The comparison with the state-of-the-art approaches, trained on HHD~\cite{belay2023hhd}. In bold, the best performant method is highlighted.}
\label{tab:sota}
\renewcommand{\arraystretch}{1.5} 

\begin{tabular}{lc|c|c|c|c}
\toprule
\multirow{2}{*}{\textbf{Method Name}} & \multirow{2}{*}{\textbf{Year}} & \multicolumn{2}{c}{\textbf{CER}} & \multicolumn{2}{c}{\textbf{NED}} \\
\cmidrule(lr){3-4} \cmidrule(lr){5-6}
 & & \textbf{IID} & \textbf{OOD} & \textbf{IID} & \textbf{OOD} \\
\midrule
Human-performance\cite{belay2023hhd} & - &25.39 &33.20 &23.78 &33.70 \\
SVTR\cite{du2022svtr} &2022 &19.78 &30.82 &17.95 &28.00 \\
ASTER\cite{shi2018aster} &2018 &24.43 &35.13 &20.88 &30.75 \\
ABINet\cite{fang2021read} &2021 &21.49 &30.82 &18.11 &28.84 \\
CRNN\cite{shi2016end} &2016 &21.04 &29.86 &21.01 &29.29 \\
Plain-CTC\cite{belay2020amharic} &2020 &20.88 &33.56 &19.09 &31.90 \\
Attn-CTC\cite{belay2021blended} &2021 &19.42 &33.07 &21.01 &32.92 \\
HPopt-Plain-CTC\cite{belay2023hhd} &2024 &19.42 &32.01 &17.77 &29.02 \\
HPopt-Attn-CTC\cite{belay2023hhd} &2024 &{16.41} &{28.65} &{16.06} &{27.37} \\
\textbf{Ours } &2024 &\textbf{15.48} &\textbf{26.95} &\textbf{15.00} &\textbf{26.50} \\
\bottomrule
\end{tabular}
\end{table}


Our method achieves state-of-the-art performance. We can see from Table~\ref{tab:sota}, that our model obtains 15.48 for CER using the IID test set, and 27.96 for the OOD test set. This not only surpasses existing OCR models presented in the table but also outperforms human-level recognition, which was used as a baseline for comparison in recognizing historical Ethiopic script in~\cite{belay2023hhd}. Figure~\ref{fig:loss} show the loss values across epochs, and shows the accuracy of classification during train time, showing that the model converges in few epochs.




\section{Conclusion}\label{sec:conclusions}
The proposed model exhibits a strong capacity for accurately and efficiently recognizing characters in the Ge'ez script. The utilization of CNNs for feature extraction, in conjunction with LSTMs, has demonstrated its effectiveness in addressing the challenges of handwritten character recognition. This paper makes a substantial contribution to the domain of document digitalization and preservation, surpassing the previous SOTA approach, specifically for manuscripts that are not adequately represented in digital format. Future research will focus on expanding the model's capability to handle more complex script scenarios and enhancing its adaptability to process novel, unfamiliar information. 
Furthermore, the future direction involves exploring tiny machine learning~\cite{capogrosso2024machine} and split computing~\cite{cunico2022split} techniques to optimize execution and improve real-world implementation of this recognition system, making it more scalable and accessible for widespread use, particularly in resource-constrained environments.


\bibliographystyle{ieeetr}
\bibliography{main.bib} 

\begin{thebibliography}{10}

\bibitem{plamondon2000online}
R.~Plamondon and S.~N. Srihari, ``Online and off-line handwriting recognition: a comprehensive survey,'' {\em IEEE Transactions on Pattern Analysis and Machine Intelligence}, vol.~22, no.~1, pp.~63--84, 2000.

\bibitem{lecun1998gradient}
Y.~LeCun, L.~Bottou, Y.~Bengio, and P.~Haffner, ``Gradient-based learning applied to document recognition,'' {\em Proceedings of the IEEE}, vol.~86, no.~11, pp.~2278--2324, 1998.

\bibitem{moges2017study}
A.~Moges, A.~Bewketu, D.~Dadi, and T.~Dibaba, ``A study on the performance of convolutional neural networks in offline handwritten amharic character recognition,'' in {\em 2017 International Conference on Advances in Computing, Communications and Informatics (ICACCI)}, pp.~1117--1123, IEEE, 2017.

\bibitem{krizhevsky2017imagenet}
A.~Krizhevsky, I.~Sutskever, and G.~E. Hinton, ``Imagenet classification with deep convolutional neural networks,'' {\em Communications of the ACM}, vol.~60, no.~6, pp.~84--90, 2017.

\bibitem{hochreiter1997long}
S.~Hochreiter and J.~Schmidhuber, ``Long short-term memory,'' {\em Neural computation}, vol.~9, no.~8, pp.~1735--1780, 1997.

\bibitem{wigington2017data}
C.~Wigington, S.~Stewart, B.~Davis, B.~Barrett, B.~Price, and S.~Cohen, ``Data augmentation for recognition of handwritten words and lines using a cnn-lstm network,'' in {\em 2017 14th IAPR international conference on document analysis and recognition (ICDAR)}, vol.~1, pp.~639--645, IEEE, 2017.

\bibitem{geez}
{Encyclopedia Britannica}, ``Ge'ez language,'' 2023.
\newblock Accessed on August 21, 2024.

\bibitem{belay2023hhd}
B.~H. Belay, I.~Guyon, T.~Mengiste, B.~Tilahun, M.~Liwicki, T.~Tegegne, and R.~Egele, ``Hhd-ethiopic a historical handwritten dataset for ethiopic ocr with baseline models and human-level performance,'' 2023.

\bibitem{belay2020amharic}
B.~Belay, T.~Habtegebrial, M.~Meshesha, M.~Liwicki, G.~Belay, and D.~Stricker, ``Amharic ocr: an end-to-end learning,'' {\em Applied Sciences}, vol.~10, no.~3, p.~1117, 2020.

\bibitem{belay2020ocr}
B.~Belay {\em et~al.}, ``Ocr for amharic scripts using deep learning,'' {\em African Journal of Information Systems}, vol.~12, no.~3, pp.~223--244, 2020.

\bibitem{du2022svtr}
Y.~Du, Z.~Chen, C.~Jia, X.~Yin, T.~Zheng, C.~Li, Y.~Du, and Y.-G. Jiang, ``Svtr: Scene text recognition with a single visual model,'' {\em arXiv preprint arXiv:2205.00159}, 2022.

\bibitem{shi2016end}
B.~Shi, X.~Bai, and C.~Yao, ``An end-to-end trainable neural network for image-based sequence recognition and its application to scene text recognition,'' {\em IEEE transactions on pattern analysis and machine intelligence}, vol.~39, no.~11, pp.~2298--2304, 2016.

\bibitem{shi2018aster}
B.~Shi, M.~Yang, X.~Wang, P.~Lyu, C.~Yao, and X.~Bai, ``Aster: An attentional scene text recognizer with flexible rectification,'' {\em IEEE transactions on pattern analysis and machine intelligence}, vol.~41, no.~9, pp.~2035--2048, 2018.

\bibitem{fang2021read}
S.~Fang, H.~Xie, Y.~Wang, Z.~Mao, and Y.~Zhang, ``Read like humans: Autonomous, bidirectional and iterative language modeling for scene text recognition,'' in {\em Proceedings of the IEEE/CVF conference on computer vision and pattern recognition}, pp.~7098--7107, 2021.

\bibitem{belay2021blended}
B.~H. Belay, T.~Habtegebrial, M.~Liwicki, G.~Belay, and D.~Stricker, ``A blended attention-ctc network architecture for amharic text-image recognition.,'' in {\em ICPRAM}, pp.~435--441, 2021.

\bibitem{capogrosso2024machine}
L.~Capogrosso, F.~Cunico, D.~S. Cheng, F.~Fummi, and M.~Cristani, ``A machine learning-oriented survey on tiny machine learning,'' {\em IEEE Access}, 2024.

\bibitem{cunico2022split}
F.~Cunico, L.~Capogrosso, F.~Setti, D.~Carra, F.~Fummi, and M.~Cristani, ``I-split: Deep network interpretability for split computing,'' in {\em 2022 26th International Conference on Pattern Recognition (ICPR)}, pp.~2575--2581, IEEE, 2022.

\end{thebibliography}
\end{document}